\documentclass{article}

\usepackage[preprint]{neurips_2026}

\usepackage[utf8]{inputenc} % allow utf-8 input
\usepackage[T1]{fontenc}    % use 8-bit T1 fonts
\usepackage{hyperref}       % hyperlinks
\usepackage{url}            % simple URL typesetting
\usepackage{booktabs}       % professional-quality tables
\usepackage{amsfonts}       % blackboard math symbols
\usepackage{nicefrac}       % compact symbols for 1/2, etc.
\usepackage{microtype}      % microtypography
\usepackage{xcolor}         % colors
\usepackage{subcaption}

\usepackage[utf8]{inputenc}
\usepackage[T1]{fontenc}
\usepackage{amsmath,amssymb,amsthm}
\usepackage{graphicx}
\usepackage{xcolor}
\usepackage{booktabs}
\usepackage{multirow}
\usepackage{array}
\usepackage{hyperref}
\usepackage{geometry}
\usepackage{microtype}
\usepackage[svgnames,table]{xcolor} % 支持 \rowcolor 和颜色名称 (必须在 colortbl 之前加载)
\usepackage{colortbl}   % 支持 \rowcolor
\usepackage{makecell}
\usepackage{capt-of}

\usepackage[most]{tcolorbox}
\usepackage{xcolor}
\usepackage{listings}
\definecolor{lightgray}{gray}{0.95}
\definecolor{deepblue}{RGB}{70,130,180}
\definecolor{deepgray}{RGB}{119,136,153}
\lstdefinestyle{prompt}{
    basicstyle=\ttfamily\fontsize{7pt}{8pt}\selectfont,
    frame=none,
    breaklines=true,         % 自动换行
    breakatwhitespace=true,  % 只在空格处换行，避免单词被切断
    backgroundcolor=\color{lightgray},
    breakatwhitespace=true,
    breakindent=0pt,
    escapeinside={(*@}{@*)},
    numbers=none,
    numbersep=5pt,
    xleftmargin=5pt,
    aboveskip=2pt,
    belowskip=2pt,
}
\tcbset{
  aibox/.style={
    top=2pt,
    bottom=2pt,
    left=2pt,
    right=2pt,
    boxsep=2pt,      % 标题与内容之间的额外间距
    colback=white,
    enhanced,
    center,
  }
}
\newtcolorbox{AIbox}[2][]{aibox, title=#2,#1}
\usepackage{pifont}
\usepackage{graphicx}

\usepackage{fontawesome5} 
\definecolor{hfdataset}{RGB}{255, 165, 0}   % 橙黄
\definecolor{hfmodel}{RGB}{106, 90, 205}     % 蓝紫

\usepackage{amsthm}

\theoremstyle{definition}
\title{ACC: Compiling Agent Trajectories for Long-Context Training}

\author{%
  Qisheng Su$^{1,2}$, Zhen Fang$^{1}$, Shiting Huang$^{1}$, Yu Zeng$^{1}$, Yiming Zhao$^{1}$, \\
  \textbf{Kou Shi$^{1}$, Ziao Zhang$^{1}$, Lin Chen$^{1}$, Zehui Chen$^{1}$, Lijun Wu$^{3}$, Feng Zhao$^{1}$\thanks{Corresponding Author}} \\
  \vspace{0.3em}
  $^{1}$MoE Key Lab of BIPC, University of Science and Technology of China \\
  $^{2}$Shanghai Innovation Institute \\
  $^{3}$Shanghai AI Laboratory \\
  \texttt{nicksu@mail.ustc.edu.cn} \quad \texttt{fzhao956@ustc.edu.cn} \\
  \textcolor{hfdataset}{\faIcon{database}}~\textbf{Dataset:}\url{https://huggingface.co/datasets/groundhogLLM/ACC-dataset} \\ \textcolor{hfmodel}{\faIcon{server}}~\textbf{Checkpoint:}\url{https://huggingface.co/groundhogLLM/ACC-Qwen3-30B-A3B}
}

\begin{document}

\maketitle

\begin{abstract}
Recent development of agents has renewed demand for long-context reasoning capacity of LLMs. However, training LLMs for this capacity requires costly long-document curation or heuristic context synthesis. We observe that agents produce massive trajectories when solving problems, invoking tools and receiving environment observations across many turns. The evidence needed to answer the original question is thus scattered throughout these turns, requiring integration of distant context segments. Nevertheless, standard agent SFT masks tool responses and only trains turn-level tool selection, creating a supervision blind spot where these scattered signals go unused. We propose Agent Context Compilation (ACC), which converts trajectories from search, software engineering, and database querying agents into long-context QA pairs that combine the original question with tool responses and environment observations gathered across multiple turns, training the model to answer directly without tool use. This makes the dependencies between the question and the evidence explicit, enabling direct supervision of long-context reasoning over distant segments without additional annotation. ACC is a simple but effective approach that can be combined with any existing long-context extension or training method, providing scalable supervised fine-tuning data. We validate ACC on long-range dependency modeling tasks through MRCR and GraphWalks, challenging benchmarks requiring cross-turn coreference resolution and graph traversal over extended contexts. Training Qwen3-30B-A3B with ACC achieves 68.3 on MRCR (+18.1) and 77.5 on GraphWalks (+7.6), results comparable to Qwen3-235B-A22B, while preserving general capabilities on GPQA, MMLU-Pro, AIME, and IFEval. Further mechanism analysis reveals that the ACC-trained model exhibits task-adaptive attention restructuring and expert specialization. Dataset and checkpoints are released publicly.
\end{abstract}
\section{Introduction}
\label{sec:intro}

Recently, the rise of agents has brought fresh attention to long-context reasoning for LLMs\cite{gpt54,opus46,gemini31,qwen35}, since agents work through many turns of tool calls and models need to handle increasingly long inputs. However, conventional training of LLMs for this capacity  relies on costly long-document curation or heuristic context synthesis. Curating annotated long documents requires precise evidence labeling and intensive quality filtering. Heuristic synthesis gathers contexts without the complex dependencies that actual problem solving creates. These limitations severely restrict scalable training for long-span reasoning and motivate the exploration of alternative supervision sources.

Agents produce massive multi-turn trajectories when solving problems, invoking tools and receiving tool responses across many turns. The evidence needed to answer the original question is scattered throughout these turns, requiring integration of distant context segments. Although these trajectories can be directly used for supervised fine-tuning, standard practice masks out tool responses and only supervises turn-level tool selection. This creates a supervision blind spot that leaves scattered evidence signals unused and severely limits the development of long-context capabilities.

\begin{figure*}[t]
    \centering
    \includegraphics[width=1\textwidth]{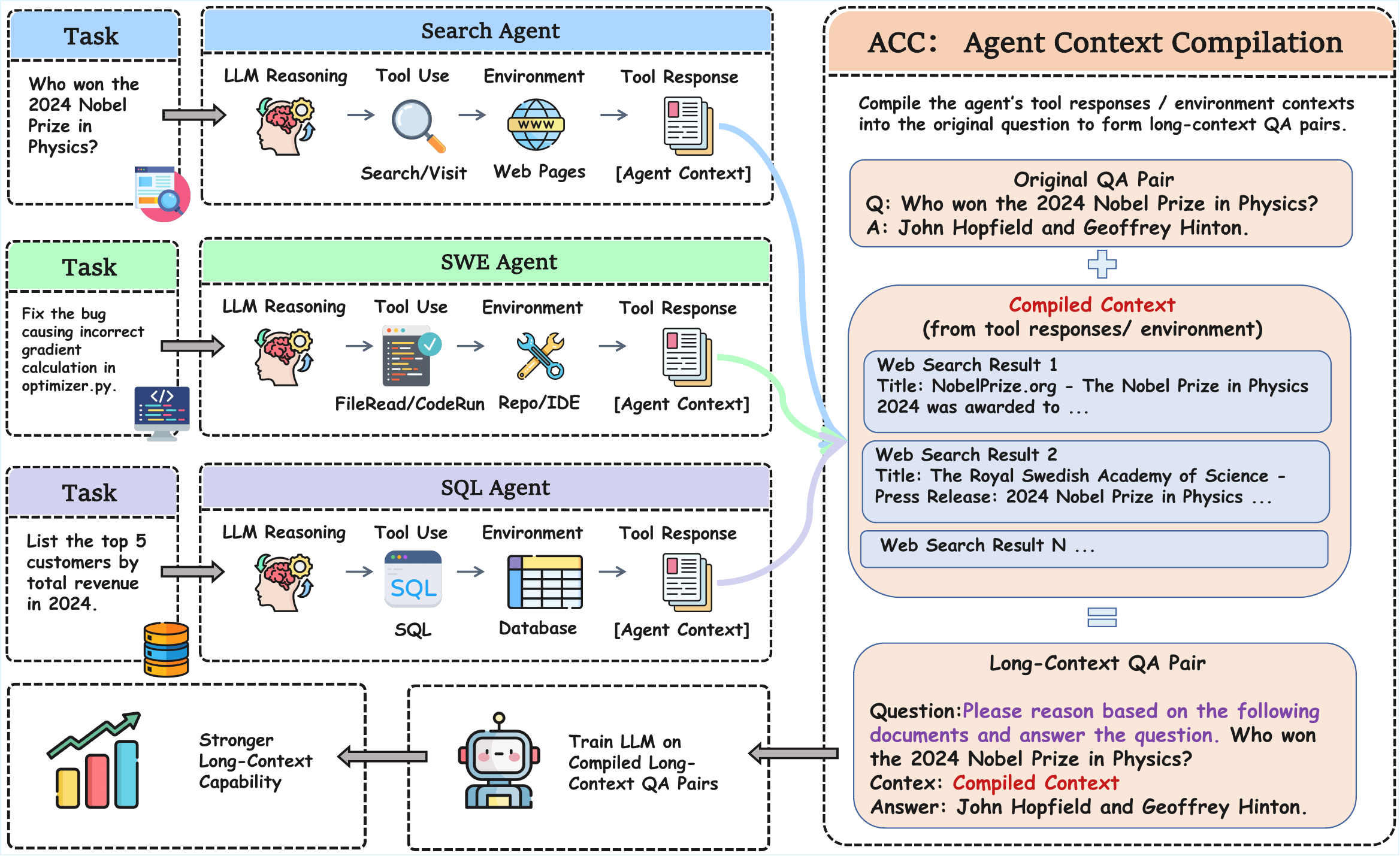}    
        \vspace{-1.5em}
\caption{\textbf{Overview of ACC.} Multi-turn agent trajectories (Search, SWE, SQL) are compiled into long-context QA pairs by assembling tool responses and environment contexts.} 
    \label{fig:overview}
    \vspace{-0.5em}
\end{figure*}

To address this, we propose \textbf{Agent Context Compilation (ACC)}, which converts agent trajectories into long-context training data without additional human annotation. By assembling the original question with tool responses and environment observations gathered across multiple turns into one context, ACC makes the dependencies between the question and scattered evidence explicit, enabling direct supervision of long-context reasoning without additional annotation. \textbf{ACC is a simple but effective approach that can be combined with existing long-context extension or training method, providing scalable supervised fine-tuning data.} Figure~\ref{fig:overview} illustrates the ACC pipeline.

We apply ACC to three representative agent classes including search agents that retrieve web pages to answer complex questions, SWE agents that inspect source files to resolve issues, and SQL agents that query relational tables for structured analytics. In each case, we compile \textbf{answer-verified trajectories} into long-context training pairs, taking the answer directly from the final output without additional human annotation.

We validate ACC on long-range dependency modeling tasks through MRCR and GraphWalks\cite{gpt41}, two particularly challenging benchmarks requiring cross-turn coreference resolution and graph traversal over extended contexts. Training Qwen3-30B-A3B with ACC achieves 68.3 on MRCR (+18.1) and 77.5 on GraphWalks (+7.6), results comparable to Qwen3-235B-A22B, while preserving general capabilities on GPQA, MMLU-Pro, AIME, and IFEval. Mechanism analysis further reveals that the ACC-trained model exhibits task-adaptive attention restructuring and expert specialization, reflecting flexible adaptation to distinct long-range reasoning demands.

\textbf{Contributions.} Our main contributions are summarized as follows. (1) We propose Agent Context Compilation (ACC), a method that converts multi-turn agent trajectories into long-context training QAs. (2) We show that the ACC-trained Qwen3-30B-A3B achieves results comparable to Qwen3-235B-A22B on long-range dependency modeling benchmarks including MRCR and GraphWalks, while preserving general capabilities. (3) Through mechanism analysis, we observe task-adaptive attention restructuring and expert specialization emerging after ACC training, suggesting that the acquired long-range capacity manifests as flexible, task-specific patterns.
\section{Related Work}
\label{sec:related}

\subsection{Long-Context Capacity Evaluation}

Evaluating long-context capabilities has evolved significantly. Early benchmarks such as NIAH \cite{niah} tested surface-level retrieval by embedding specific facts within distractor text. RULER \cite{ruler} extended this with variable tracking, aggregation, and multi-hop reasoning tasks. LongBench \cite{longbenchv2} introduced diverse real-world tasks including QA, summarization, and code understanding. However, performance on these benchmarks has largely saturated, as they mainly test localized retrieval or single-turn reasoning within long contexts. Classic benchmarks such as Musique \cite{musique} and NarrativeQA \cite{narrativeqa} further targeted multi-hop reasoning and long-document narrative understanding. More recently, OpenAI released MRCR (Multi-Round Coreference Resolution) and GraphWalks \cite{gpt41} as direct tests of long-range dependency modeling. By requiring cross-turn coreference resolution and graph traversal over extended contexts, they are substantially harder than prior single-turn or retrieval tasks, and have become standard benchmarks for mainstream large model releases.

\subsection{Long-Context Extension and Training}

Recent efforts to improve long-context capabilities generally fall into four categories. First, pre-training methods modify position embeddings or attention mechanisms. MrRoPe \cite{mrrope} applies RoPE interpolation and NTK-aware frequency scaling to broaden the context window. ROPE++ \cite{rope++} reuses the discarded imaginary component of RoPE's complex form to build parallel attention heads for improved length extrapolation. Native Sparse Attention \cite{nsa} and Mamba-3 \cite{mamba3} reduce complexity through sparse and linear attention. Second, some works focuses on constructing high-quality long documents for pre-training data. Longwanjuan \cite{longwanjuan} filters texts by coherence, cohesion, and complexity. LiteLong \cite{litelong} leverages book taxonomies and multi-agent debate for corpora retrieval and concatenation. Quest \cite{quest} predicts possible questions and clusters core keywords to stitch short documents. These methods synthesize long texts rather than post-training QA pairs. Third, post-training recipes combine synthetic data with RL. longRLVR \cite{longrlvr} generates QA pairs with precise evidence block annotations from long texts. LongPO \cite{longpo} extracts key short chunks to build short-long preference pairs and applies short-to-long KL constraints in DPO. LoongRL \cite{loongrl} proposes KeyChain to insert irrelevant documents for hard long-context synthesis and stabilizes GRPO with rule rewards and no entropy term. Fourth, employ agent frameworks at inference time to manage long-context memory. QwenLong-L1.5 \cite{qwenlongl1.5} cleans multi-source documents, builds knowledge graphs, and applies AEPO for dynamic entropy control. MemAgent \cite{memagent} mixes irrelevant HotpotQA documents and uses Multi-Conv DAPO to decompose long questions into multi independent conversations with memory updates. Our work differs by using agent trajectories as a direct data source for long-context reasoning training, rather than modifying architectures, synthesizing pre-training documents, or relying on complex post-training RL pipelines.

\section{Method}
\label{sec:method}

\subsection{The Supervision Blind Spot of Agent SFT}
\label{sec:blind_spot}

Standard agent SFT masks all tool responses (observations) and only supervises turn-level reasoning and actions. The model therefore never learns to integrate evidence scattered across multiple turns.

An agent trajectory consists of $k-1$ interaction turns followed by a final answer turn
\[
\tau=(q,(r_1,a_1,o_1),\dots,(r_{k-1},a_{k-1},o_{k-1}),(r_k,y)),
\]
where $r_t$ is reasoning, $a_t$ is action, $o_t$ is tool response (observation), and $(r_k,y)$ is the final reasoning-answer pair. The history up to turn $t$ is $\mathcal{H}_{<t}=(r_1,a_1,o_1,\ldots,r_{t-1},a_{t-1},o_{t-1})$\footnote{We present interleaved reasoning traces for clarity. Non-interleaved variants do not affect our conclusions.}. Tool responses are masked from the loss and only model-generated tokens are supervised.

Formally, the standard objective is
\begin{equation}
    \mathcal{L}_{\text{agent}} = -\sum_{t=1}^{k}\sum_{j\in\mathcal{I}_t} \log P(\text{token}_j \mid \mathcal{H}_{<t}, \text{token}_{<j}),
    \label{eq:agent_loss}
\end{equation}
where $\mathcal{I}_t=r_t\cup a_t$ for $t<k$ and $\mathcal{I}_k=r_k\cup y$.

Grouping Eq.~\eqref{eq:agent_loss} by turn reveals its structure
\begin{equation}
    \resizebox{\textwidth}{!}{$
    \mathcal{L}_{\text{agent}} 
    = \underbrace{\sum_{t=1}^{k-1}\Bigl( -\sum_{j\in r_t\cup a_t} \log P(\text{token}_j \mid \mathcal{H}_{<t}, \text{token}_{<j}) \Bigr)}_{\text{local next-tool selection}}
    \;+\;
    \underbrace{\Bigl( -\sum_{j\in r_k\cup y} \log P(\text{token}_j \mid \mathcal{H}_{<k}, \text{token}_{<j}) \Bigr)}_{\text{final answer prediction}}.
    $}
    \label{eq:agent_loss_decomposed}
\end{equation}

The first $k-1$ terms supervise only local reasoning and tool selection at each turn. Consider a token in tool response $o_t$ at turn $t<k$. Excluded from the loss, it receives gradients only indirectly through subsequent unmasked tokens. The dominant signal flows along a \textbf{short path} to the next action $a_{t+1}$, where $o_t$ lies in the immediate context. Any gradient relevant to the final answer $y$ must back-propagate through a \textbf{long chain} of intermediate turns to reach $o_t$, and is heavily weakened. Consequently, these intermediate turns act as a \textbf{supervision filter}, so $o_t$ is updated primarily to support local action prediction, ignoring answer-relevant features unless they also serve local needs. This is the supervision blind spot of agent SFT.

\subsection{Agent Context Compilation}
\label{sec:acc}

ACC solves this problem by gathering all evidence into one long context $C$ and training the model to write a reasoning trace $r$ and final answer $y$ directly from the question $q$ and context $C$. The new training objective is
\begin{equation}
\mathcal{L}_{\text{ACC}} = -\sum_{j\in r\cup y} \log P(\text{token}_j \mid q, C, \text{token}_{<j}).
\label{eq:acc_loss}
\end{equation}

Unlike Eq.~\eqref{eq:agent_loss}, this objective contains no intermediate action terms, so the final answer supervision reaches every evidence token directly without being filtered through turn-level tool selection. The model therefore learns to integrate scattered evidence into a global answer instead of merely optimizing local next-tool selection.

Given a set of \textbf{answer-verified trajectories} $\mathcal{T}=\{\tau_1,\dots,\tau_N\}$, ACC converts each trajectory into a training example
\[
\tau_i \mapsto (x_i, y_i, r_i),
\]
producing a dataset $\mathcal{D}=\{(x_i,y_i,r_i)\}_{i=1}^M$. Here $x_i=(q_i,C_i)$ combines the original query with the compiled context, $y_i$ is the final answer from the trajectory, and $r_i$ is its reasoning trace.

\subsection{Context Construction}
\label{sec:context_construction}

For each trajectory we extract structured evidence pieces $\text{Evi}(\tau)=[e_1,\dots,e_m]$ such that the aggregated context alone suffices to answer $q$ without tool use. For \textbf{search trajectories} we extract the full text of visited pages and include unvisited candidate results as distractors. For \textbf{SWE trajectories} we extract files involved in the correct patch and include additional context files inspected during debugging as distractors. For \textbf{SQL trajectories} we extract the complete contents of all tables queried during the trajectory.

To increase task difficulty, we apply a random permutation $\pi$ over $\{1,\dots,m\}$ and concatenate the pieces into a compiled context
\begin{equation}
    C_i = \operatorname{Concat}(e_{\pi(1)}, e_{\pi(2)}, \dots, e_{\pi(m)}), \quad \text{with } |C_i| \leq B,
    \label{eq:context_compile}
\end{equation}
where $B$ is the token budget. Because evidence pieces are self-contained, shuffling forces the model to locate relevant information via semantic association rather than sequential position.

\textbf{Answer-verified trajectories} contain correct answers but lack explicit reasoning traces. We employ \textbf{DeepSeek-V3.2-Thinking} to generate candidate rationales and retain only those that lead to the correct answer $y_i$. For Search and SQL, these traces are obtained via direct rollout, with pass rates near 100\% and 50\%, respectively. 
For SWE, direct rollout achieves only near 10\% accuracy, which is too low to scale. We therefore synthesize answer-conditioned reasoning traces from the compiled context and verified patch instead, with all outputs retained (100\% yield, see Appendix~\ref{app:swe_rationale}).  The final training example is the triple $(x_i,y_i,r_i)$, where $x_i=(q_i,C_i)$ and $r_i$ is the retained reasoning trace.

\vspace{-0.5em}
\begin{figure}[!ht]
\begin{AIbox}{Search Agent Trajectory Compilation Example}
{\color{deepblue}\bf \normalsize Original Question-Answer Pair:}

\textbf{Question}: Which track, credited as a cover of the French new‑wave group, appears as a European‑bonus track on a French symphonic black‑metal band's early‑2000s album, is featured as a roughly five‑minute live performance on that group's mid‑1990s live album recorded for radio, and is listed as the third song on Disc~One of the group's late‑1990s live collection?

\textbf{Groundtruth}: Les Tzars
\tcblower
{\color{deepblue}\bf \normalsize Original Agentic Trajectory (abridged):} 
\begin{lstlisting}[style=prompt, escapeinside={(*@}{@*)}]
Turn 1 (Search): "French symphonic black metal Indochine 
                 cover European bonus track"
Result: (*@\textcolor{blue}{\textbf{[Doc A]}}@*) Anorexia Nervosa - Redemption Process... 
        8. Les Tzars (Indochine Cover)... limited press...
        [Doc B] Indochine is a French new wave band...
        [Doc C] Eating disorders treatment and recovery...
-> (Visit) (*@\textcolor{blue}{\textbf{Doc A}}@*)

Turn 2 (Search): "Indochine live album 1994 radio Les Tzars"
Result: (*@\textcolor{blue}{\textbf{[Doc D]}}@*) Radio Indochine is the second live album... 
        released in 1994... Les tzars - 5:12...
        [Doc E] Indochine au Zenith 1986 live album...
        [Doc F] Neuropsychological assessment of ED...
-> (Visit) (*@\textcolor{blue}{\textbf{Doc D}}@*)

Turn 3 (Search): "Indochine Indo Live 1997 Disc One track"
Result: (*@\textcolor{blue}{\textbf{[Doc G]}}@*) INDO LIVE is the third live album... 
        released in 1997... Disc One... Les tzars...
        [Doc H] Indochine 13 album 2017...
        [Doc I] Hiroshima mon amour 1959 film...
-> (Visit) (*@\textcolor{blue}{\textbf{Doc G}}@*)

Turn 4 (Answer): Les Tzars (*@\textcolor{ForestGreen}{\ding{51}}@*)
\end{lstlisting}

\vspace{0.5em}
{\color{deepblue}\bf \normalsize ACC-Compiled Question-Answer Pair (abridged):}
\begin{lstlisting}[style=prompt, escapeinside={(*@}{@*)}]
Question: Please reason based on the following documents and answer the question.
Which track, credited as a cover of the French new-wave group, appears as a European-bonus track...

Documents:
(*@\textcolor{blue}{\textbf{[Doc G]}}@*) INDO LIVE is the third live album by French new wave 
band Indochine... released in 1997... Disc One... Les tzars...
(*@\textcolor{red}{\textbf{[Doc C]}}@*) Eating disorders treatment and recovery... (*@\textcolor{red}{\textit{(distractor)}}@*)
(*@\textcolor{blue}{\textbf{[Doc A]}}@*) Anorexia Nervosa - Redemption Process... 8. Les Tzars 
(Indochine Cover)... limited press of six thousand copies...
(*@\textcolor{blue}{\textbf{[Doc D]}}@*) Radio Indochine is the second live album by French new 
wave band Indochine... released in 1994... Les tzars - 5:12...


Answer: Les Tzars
\end{lstlisting}
\end{AIbox} 
\caption{\textbf{Search Agent Trajectory Compilation Example.} The top section shows the original question and ground truth answer. The middle section shows the original agentic trajectory (documents visited are highlighted in \textcolor{blue}{blue}, documents returned by search but never visited are highlighted in \textcolor{red}{red}). The bottom section shows the ACC compiled QA. Examples for SWE and SQL agents are provided in Appendix~\ref{app:examples}.}
\label{fig:search_example}
\vspace{-1em}
\end{figure}

\section{Experiments}
\label{sec:experiments}

\subsection{Experimental Setup}
\label{sec:setup}

\textbf{Base Model.} We use Qwen3-30B-A3B-Thinking \cite{yang2025qwen3technicalreport} as our base model.

\begin{figure}[t]
    \centering
    % Left: Token distribution figure
    \begin{minipage}[t]{0.55\linewidth}
        \vspace{0pt}
        \centering
        \includegraphics[width=\linewidth]{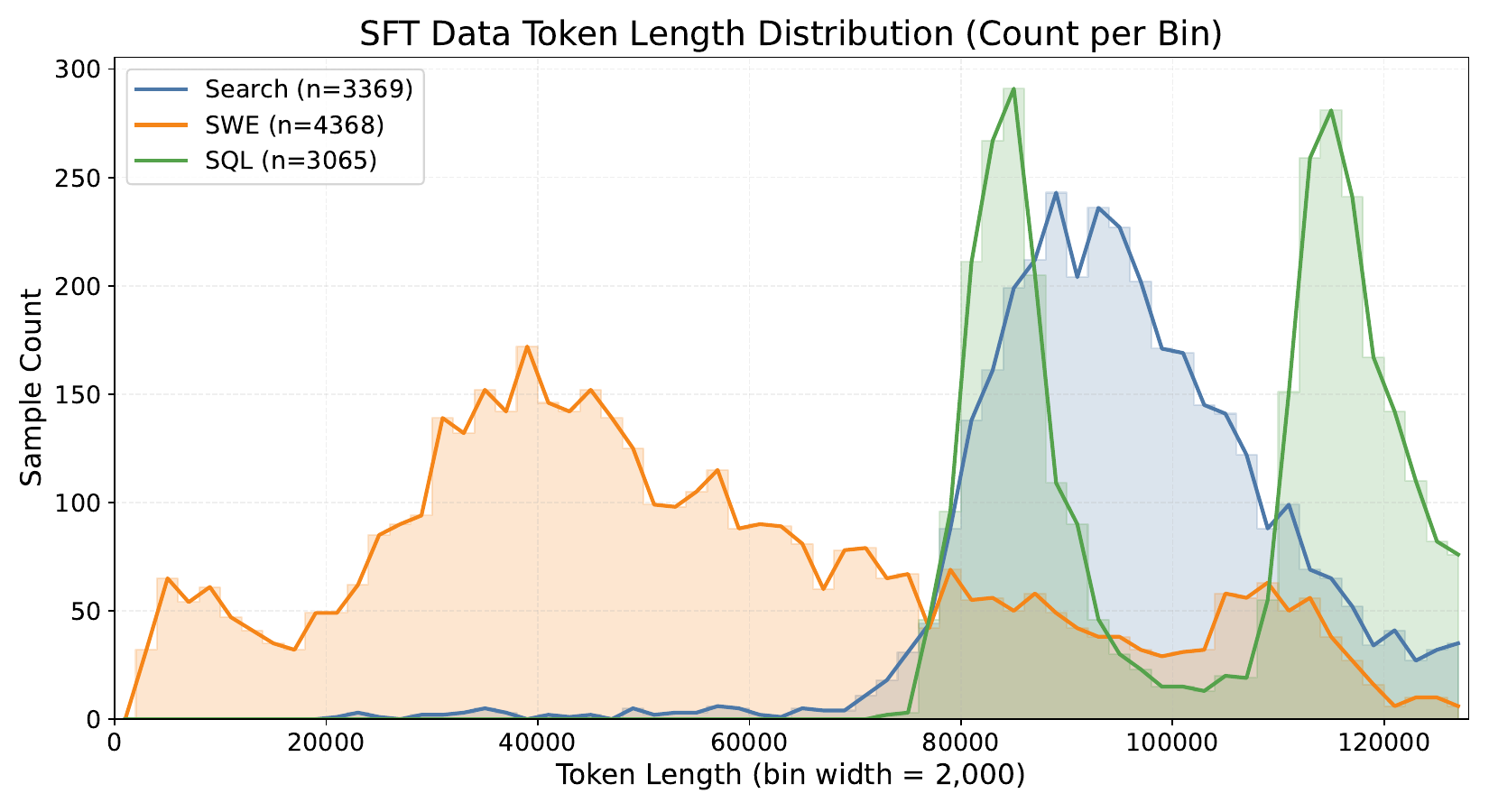}
        \vspace{-1.2em}
        \caption{\textbf{Token length distribution of the ACC training data.} We bin the samples by token count and plot the per-bin frequency for the training data compiled from each agent type.}
        \label{fig:datalength}
    \end{minipage}
    \hfill
    % Right: Training config table
    \begin{minipage}[t]{0.42\linewidth}
        \vspace{0pt}
        \centering
        \begingroup
        \footnotesize
        \setlength{\tabcolsep}{3pt}
        \captionof{table}{Training parameters for our supervised fine-tuning.}
        \label{tab:training_config}
        \vspace{0.3em}
        \begin{tabular}{@{}ll@{}}
            \toprule
            \textbf{Hyperparameter} & \textbf{Value} \\
            \midrule
            Sequence length & 131,072 tokens \\
            Global batch size & 16 \\
            Learning rate & $1\times10^{-5}$ (min $1\times10^{-6}$) \\
            LR schedule & Cosine with 5\% warmup \\
            Optimizer & \begin{tabular}[t]{@{}l@{}} AdamW \\ ($\beta_1{=}0.9$, $\beta_2{=}0.999$, \\ weight decay 0.1) \end{tabular} \\
            Loss & \begin{tabular}[t]{@{}l@{}} Cross-entropy \\ (chunk size 1024) \end{tabular} \\
            Sequence parallelism & 8 \\
            Expert parallelism & 1 \\
            Training epochs & 4 \\
            \bottomrule
        \end{tabular}
        \endgroup
    \end{minipage}
    \vspace{-0.5em}
\end{figure}

\textbf{Training Configuration.} We compile 10,802 trajectories in total (Search: 3,369; SWE: 4,368; SQL: 3,065), with compiled context lengths ranging from 2K to 128K tokens and distinct per-agent length distributions (Figure \ref{fig:datalength}). The details of training parameters are summarized in Table \ref{tab:training_config}.

\textbf{Evaluation Benchmarks.} We primarily evaluate on long-range dependency modeling benchmarks including MRCR~\cite{gpt41} (multi-round coreference resolution) and GraphWalks~\cite{gpt41} (graph traversal), which require tracking long-range relational dependencies across extended contexts. We also monitor general capabilities on GPQA-Diamond~\cite{gpqa}, MMLU-Pro~\cite{mmlu}, AIME~\cite{aime}, and IFEval~\cite{ifeval} to check for negative transfer.

\subsection{Main Results}

Table~\ref{tab:main} presents our main results on long-range dependency modeling benchmarks. On MRCR, ACC improves both the 2-needle and 4-needle settings, yielding an overall score of 68.28 (+18.09). On GraphWalks, ACC improves both the Parents and BFS sub-tasks, yielding an overall precision of 77.51 (+7.59). These results are comparable to Qwen3-235B-A22B on these long-range dependency modeling benchmarks despite having nearly 8$\times$ fewer active parameters. For completeness, we also report results on additional long-context benchmarks in Appendix~\ref{app:extended_results}.

\begin{table}[t]
\centering
\caption{Long-range dependency modeling benchmark results (avg@3). Numbers in parentheses show improvement over the Qwen3-30B-A3B-Thinking baseline. All results use default inference configurations without manual tuning of reasoning effort.}
\label{tab:main}
\footnotesize
\setlength{\tabcolsep}{4pt}
\newcommand{\res}[2]{%
  \begin{tabular}[c]{@{}c@{}}\textbf{#1}\\[-2pt]\textcolor{red}{\scriptsize(+#2)}\end{tabular}%
}
\newcommand{\ressub}[2]{%
  \begin{tabular}[c]{@{}c@{}}#1\\[-2pt]\textcolor{red}{\scriptsize(+#2)}\end{tabular}%
}
\begin{tabular}{l ccc ccc}
\toprule
\multirow{2}{*}{\textbf{Model}} & \multicolumn{3}{c}{\textbf{MRCR}\textsuperscript{1}} & \multicolumn{3}{c}{\textbf{GraphWalks}\textsuperscript{2}} \\
\cmidrule(lr){2-4} \cmidrule(lr){5-7}
& \textbf{2-needle} & \textbf{4-needle} & \textbf{Overall} & \textbf{Parents} & \textbf{BFS} & \textbf{Overall} \\
\midrule
\rowcolor{Honeydew}
\multicolumn{7}{l}{\textit{\textbf{Base Model}}} \\
Qwen3-30B-A3B-Thinking & 61.84 & 38.41 & 50.19 & 71.19 & 68.47 & 69.92 \\
\midrule
\rowcolor{AliceBlue}
\multicolumn{7}{l}{\textit{\textbf{Our Method}}} \\
\rowcolor{AliceBlue!15}
\textbf{Qwen3-30B-A3B-Thinking + ACC (Ours)} & 
  \ressub{76.90}{15.06} & 
  \ressub{59.57}{21.16} & 
  \res{68.28}{18.09} & 
  \ressub{81.50}{10.31} & 
  \ressub{72.95}{4.48} & 
  \res{77.51}{7.59} \\
\midrule
\rowcolor{MistyRose!40}
\multicolumn{7}{l}{\textit{\textbf{Strong Baselines}}} \\
Qwen3-235B-A22B-Thinking & 74.98 & 59.96 & 67.51 & 78.53 & 74.45 & 76.63 \\
GPT-OSS-120B & 46.72 & 29.16 & 38.00\textsuperscript{3} & 75.92 & 61.82 & 69.34 \\
DeepSeek-V3.2-Thinking & 81.60 & 60.32 & 71.01 & 89.87 & 80.26 & 85.39 \\
GPT-5.1-Thinking & 73.75 & 47.93 & 60.91 & 81.41 & 29.41 & 57.14 \\
GLM-4.6-Thinking & 73.63 & 55.33 & 64.53 & 80.29 & 77.41 & 78.95 \\
Kimi-K2-Thinking & 68.01 & 47.96 & 58.05 & 84.34 & 75.17 & 80.04 \\
\bottomrule
\end{tabular}
\\[4pt]
\begin{minipage}{\linewidth}
\footnotesize\raggedright
\textsuperscript{1}MRCR reports overall and sub-task (2-needle, 4-needle) scores, following the evaluation setting in Qwen-Long-L1.5~\cite{qwenlongl1.5}.\\
\textsuperscript{2}GraphWalks reports overall and sub-task (Parents, BFS) precision, following the evaluation setting in LongCat-Flash-Omni~\cite{meituanlongcatteam2025longcatflashomnitechnicalreport}.\\
\textsuperscript{3}GPT-OSS-120B was evaluated using the HuggingFace checkpoint via vLLM, and its lower MRCR scores largely result from frequent harmony-format parsing failures when processing multi-turn inputs.

\end{minipage}
\end{table}

\subsection{General Capability Preservation}

Long-context training often raises concerns about negative transfer to general capabilities. As shown in Table~\ref{tab:general}, our ACC-trained model achieves slight improvements on GPQA-Diamond (+2.49), MMLU-Pro (+1.50) and AIME'25(+3.33), while performance on AIME'24 and IFEval remains stable. These results suggest that ACC does not introduce noticeable degradation to general abilities.

\begin{table}[htbp]
\centering
\caption{General capability evaluation (avg@3). No significant negative transfer is observed.}
\label{tab:general}
\footnotesize
\setlength{\tabcolsep}{5pt}
\newcommand{\res}[3]{%
  \begin{tabular}[c]{@{}c@{}}#1\\[-2pt]\textcolor{#3}{\scriptsize(#2)}\end{tabular}%
}
\resizebox{\linewidth}{!}{%
\begin{tabular}{lccccc}
\toprule
\textbf{Model} & \textbf{GPQA-Diamond} & \textbf{MMLU-Pro} & \textbf{AIME'24} & \textbf{AIME'25} & \textbf{IFEval} \\
\midrule
\rowcolor{Honeydew}
\multicolumn{6}{l}{\textit{\textbf{Base Model}}} \\
Qwen3-30B-A3B-Thinking & 67.71 & 74.50 & 90.00 & 86.67 & 86.69 \\
\midrule
\rowcolor{AliceBlue}
\multicolumn{6}{l}{\textit{\textbf{Our Method}}} \\
\rowcolor{AliceBlue!15}
\textbf{Qwen3-30B-A3B-Thinking + ACC (Ours)} & 
  \res{\textbf{70.20}}{+2.49}{red} & 
  \res{\textbf{76.00}}{+1.50}{red} & 
  \res{90.00}{0.00}{gray} & 
  \res{\textbf{90.00}}{+3.33}{red} & 
  \res{86.14}{$-$0.55}{green} \\
\bottomrule
\end{tabular}%
}
\end{table}

\begin{figure}[t]
    \centering
    % Left: UMAP figure
    \begin{minipage}[c]{0.55\linewidth}
        \vspace{0pt}  % 关键：强制顶部对齐生效
        \centering
        \includegraphics[width=\linewidth]{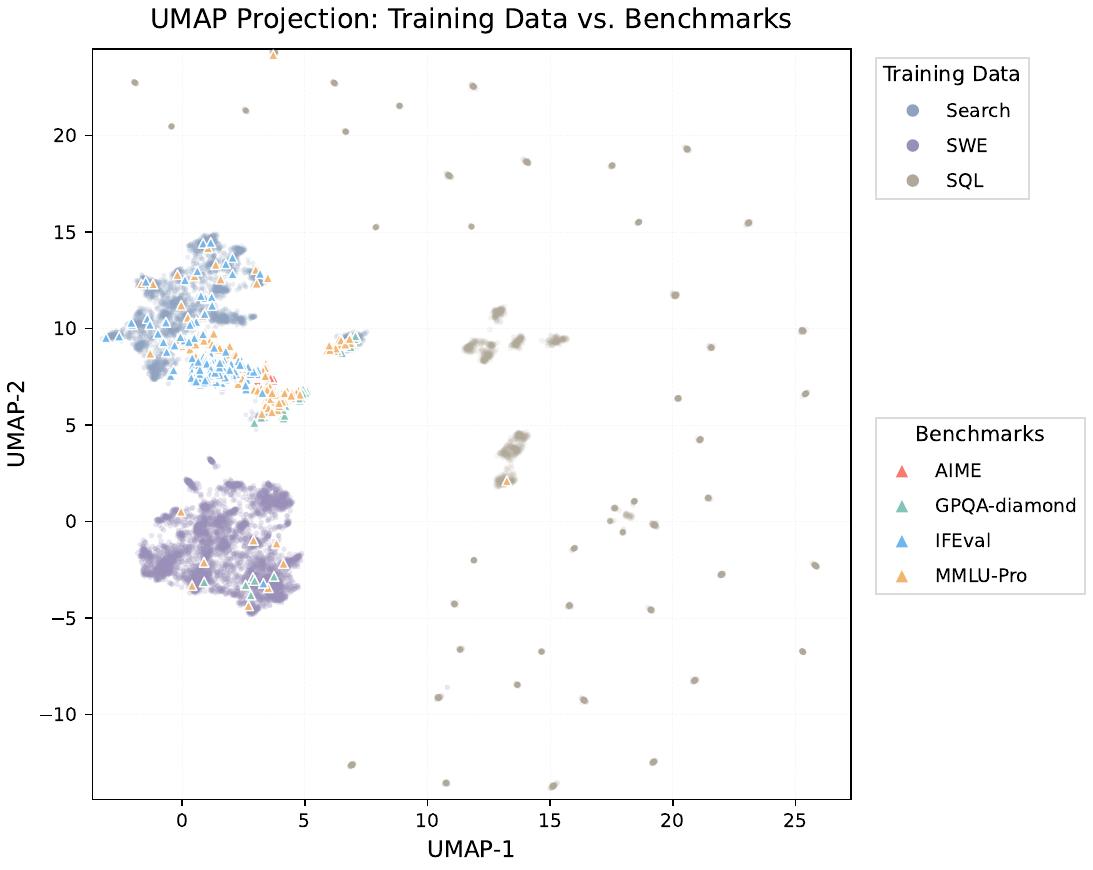}
        \captionsetup{font=footnotesize}
        \caption{Two-dimensional UMAP projection of training queries (Search, SWE, SQL) and evaluation benchmark questions. Both training and evaluation samples are represented by their question text only.}
        \label{fig:umap}
    \end{minipage}
    % \hfill
    \hspace{0.02\linewidth}
    % Right: Quantitative table
    \begin{minipage}[c]{0.40\linewidth}
        \vspace{0pt}  % 关键：强制顶部对齐生效
        \centering
        \begingroup
        \footnotesize
        \setlength{\tabcolsep}{4pt}
        \captionsetup{font=footnotesize}
        \captionof{table}{Quantitative separation between training queries and benchmark questions. Lower nearest-neighbor similarity and higher center distance both indicate limited overlap.}
        \label{tab:decontamination}
        \vspace{0.3em}
        \begin{tabular}{lcc}
            \toprule
            \textbf{Benchmark} & \textbf{NN Sim.} & \textbf{Center Dist.} \\
            \midrule
            AIME & 0.2832 & 0.8701 \\
            GPQA-Diamond & 0.3557 & 0.7150 \\
            MMLU-Pro & 0.3216 & 0.7685 \\
            IFEval & 0.3425 & 0.9216 \\
            \midrule
            \multicolumn{3}{c}{Overall AUC = 0.9986} \\
            \bottomrule
        \end{tabular}
        \endgroup
    \end{minipage}
    \vspace{-0.5em}
\end{figure}

To verify that these gains do not reflect test-set leakage, we compare the semantic distribution of training queries against benchmark questions. For each trajectory, we extract only the user question, stripping retrieved documents, code files, and database tables. Benchmark questions are similarly cleaned. Figure~\ref{fig:umap} shows the UMAP projection, and Table~\ref{tab:decontamination} reports quantitative metrics. Full details are in Appendix~\ref{app:decontamination}.

%The UMAP projection shows that Search subset queries partially overlap with general-knowledge benchmarks in the upper-left region, while SWE subset queries form a separate cluster and SQL subset queries spread broadly. This overlap is expected since our multi-hop Search queries are synthesized from Wikipedia corpora, which naturally share topical vocabulary with science and math benchmarks. Quantitative analysis suggests this is domain-level overlap rather than instance duplication. The average nearest-neighbor cosine similarity remains below 0.36, and a linear classifier achieves an AUC of 0.9986 in separating training subset queries from benchmark questions. These metrics indicate that the preserved general capabilities arise from transferable reasoning patterns rather than data leakage.

The Search subset partially overlaps with general-knowledge benchmarks. Our multi-hop Search queries are synthesized from Wikipedia corpora, which naturally share topical vocabulary with knowledge benchmarks. The SWE and SQL subsets form distinct clusters. Quantitative analysis confirms this is domain-level overlap rather than instance duplication. The average nearest-neighbor cosine similarity remains below 0.36, and a linear classifier achieves an AUC of 0.9986 in separating training queries from benchmark questions. These patterns suggest the gains reflect transferable reasoning rather than data leakage.

\subsection{Comparison with Long-Context Post-Training Methods}
\label{sec:compare_longctx}

Table~\ref{tab:compare_longctx} compares ACC with recent long-context post-training methods. QwenLong-L1.5~\cite{qwenlongl1.5} leads on MRCR through a multi-stage pipeline involving document cleaning, knowledge-graph construction, and RL. ACC surpasses it on GraphWalks while requiring only standard SFT. LongPO~\cite{longpo} and LongRLVR~\cite{longrlvr} release models trained on the Qwen2.5 and are listed for reference.\footnote{LoongRL~\cite{loongrl} does not release trained checkpoints, so we do not include it in the comparison.}

\noindent
\begin{minipage}[t]{0.48\textwidth}
\centering
\captionof{table}{Comparison with long-context post-training methods.}
\label{tab:compare_longctx}
% \small
\footnotesize
\setlength{\tabcolsep}{4pt}
\newcommand{\res}[3]{%
  \begin{tabular}[c]{@{}c@{}}#1\\[-2pt]\textcolor{#3}{\scriptsize(#2)}\end{tabular}%
}
% \resizebox{\linewidth}{!}{%
\begin{tabular}{@{}lcc@{}}
\toprule
\textbf{Model} & \textbf{MRCR} & \textbf{GraphWalks} \\
\midrule
\rowcolor{Honeydew}
\multicolumn{3}{@{}l}{\textit{\textbf{Base Model}}} \\
Qwen3-30B-A3B-Thinking & 50.19 & 69.92 \\
\midrule
\rowcolor{MistyRose!40}
\multicolumn{3}{@{}l}{\textit{\textbf{Comparison Methods}}} \\
QwenLong-L1.5-30B\textsuperscript{1} & 92.30 & 73.85 \\
Qwen2.5-7B-LongRLVR & 19.76 & 15.72 \\
Qwen2.5-14B-LongRLVR & 20.06 & 22.78 \\
Qwen2.5-7B-LongPO-128K\textsuperscript{2} & 31.50 & 12.97 \\
\midrule
\rowcolor{AliceBlue}
\multicolumn{3}{@{}l}{\textit{\textbf{Our Method}}} \\
\rowcolor{AliceBlue!15}
\textbf{+ ACC (Ours)} & \textbf{68.28} & \textbf{77.51} \\
\bottomrule
\end{tabular}%
% }
\\[4pt]
\begin{minipage}{\linewidth}
\footnotesize\raggedright
\textsuperscript{1}QwenLong-L1.5 is trained with an agent framework that is not publicly available, so we evaluate it with standard inference for contexts within 256K.\\
\textsuperscript{2}LongPO checkpoint supports up to 128K context, and test instances exceeding this limit are excluded from evaluation.
\end{minipage}
\end{minipage}%
\hfill
\begin{minipage}[t]{0.48\textwidth}
\centering
\captionof{table}{Agent-type and distractor ablations.}
\label{tab:ablation}
% \small
\footnotesize
\setlength{\tabcolsep}{4pt}
\newcommand{\res}[3]{%
  \begin{tabular}[c]{@{}c@{}}#1\\[-2pt]\textcolor{#3}{\scriptsize(#2)}\end{tabular}%
}
% \resizebox{\linewidth}{!}{%
\begin{tabular}{@{}lcc@{}}
\toprule
\textbf{Training Data} & \textbf{MRCR} & \textbf{GraphWalks} \\
\midrule
\rowcolor{Honeydew}
\multicolumn{3}{@{}l}{\textit{\textbf{Base Model}}} \\
Qwen3-30B-A3B-Thinking & 50.19 & 69.92 \\
\midrule
\rowcolor{MistyRose!40}
\multicolumn{3}{@{}l}{\textit{\textbf{Ablations}}} \\
+ Search (Agent SFT) & \res{42.16}{-8.03}{green} & \res{57.87}{-12.05}{green} \\
+ Search & \res{58.33}{+8.14}{red} & \res{44.75}{-25.17}{green} \\
+ Search (w/o distractor) & \res{54.99}{+4.80}{red} & \res{58.46}{-11.46}{green} \\
+ SWE & \res{54.82}{+4.63}{red} & \res{50.66}{-19.26}{green} \\
+ SWE (w/o distractor) & \res{51.01}{+0.82}{red} & \res{52.88}{-17.04}{green} \\
+ SQL & \res{56.44}{+6.25}{red} & \res{75.50}{+5.58}{red} \\
\midrule
\rowcolor{AliceBlue}
\multicolumn{3}{@{}l}{\textit{\textbf{Our Method}}} \\
\rowcolor{AliceBlue!15}
\textbf{+ ACC (Ours)} & \res{\textbf{68.28}}{+18.09}{red} & \res{\textbf{77.51}}{+7.59}{red} \\
\bottomrule
\end{tabular}%
% }
\end{minipage}

\subsection{Ablation Study}
\label{sec:agent_ablation}

\paragraph{Agent-type ablation.}
Raw search trajectories with Agent SFT (observations masked) underperform the base model, confirming the \textbf{supervision blind spot} in Section \ref{sec:blind_spot}. As shown in Table~\ref{tab:ablation}, we ablate ACC by training on each agent type separately. All single-agent variants improve over the baseline on MRCR (Search +8.14, SWE +4.63, SQL +6.25), indicating that compiling scattered evidence into a single context alone improves cross-turn coreference resolution. On GraphWalks, however, only SQL improves (+5.58), while Search and SWE fall behind. This gap likely reflects differences in evidence structure. SQL tables are inherently relational and suit graph traversal, whereas web pages and source files are longer continuous passages that make discrete node-level reasoning harder to learn. The full mixture surpasses all single-agent variants, showing that diverse trajectory types offer complementary coverage.

\paragraph{Distractor ablation.} 
Removing distractors from Search and SWE lowers MRCR by 3.34 and 3.81 points, confirming that including unvisited results and unopened files in the compiled context helps the model to learn localizing critical evidence. On GraphWalks, the single-agent setting shows the opposite trend, with Search and SWE without distractors gaining +13.71 and +2.22 respectively. This is because Search and SWE distractors are semantically unrelated to the query, helping the model learn noise filtering but offering little benefit for graph traversal. The full mixture, enriched by SQL's relational data, benefits from distractors for localization while preserving graph-walking capability. The full ACC mixture still achieves the best overall result (77.51).

\subsection{Mechanism Analysis}
\label{sec:capability_analysis}

To understand how ACC improves long-range dependency modeling capacity, we visualize attention distance distributions and expert routing patterns on GraphWalks and MRCR examples.

\begin{figure}[t]
    \centering
    \begin{subfigure}[b]{0.48\textwidth}
        \centering
        \includegraphics[width=\textwidth,height=0.28\textheight,keepaspectratio]{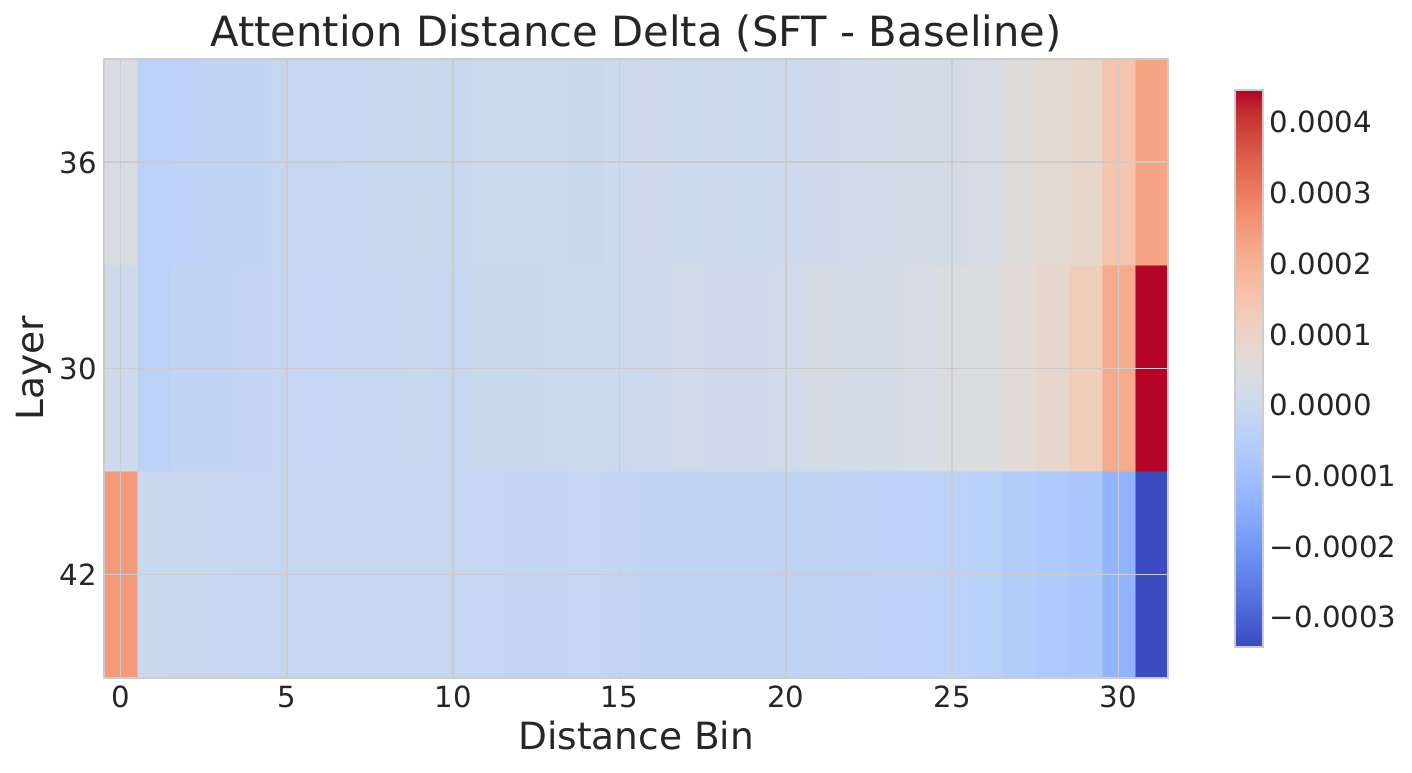}
        \caption{GraphWalks}
        \label{fig:attn_gw}
    \end{subfigure}
    \hfill
    \begin{subfigure}[b]{0.48\textwidth}
        \centering
        \includegraphics[width=\textwidth,height=0.28\textheight,keepaspectratio]{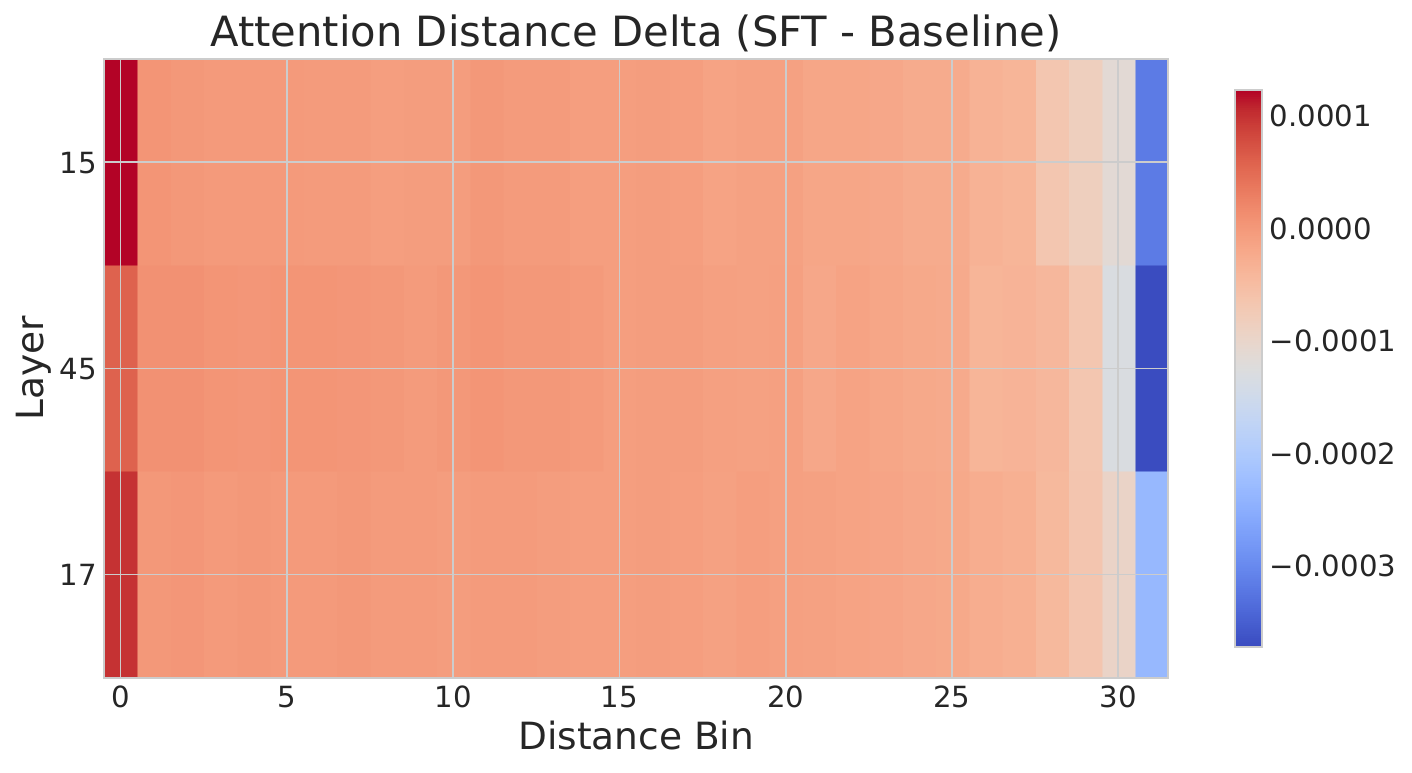}
        \caption{MRCR}
        \label{fig:attn_mrcr}
    \end{subfigure}
    \vspace{2pt}  % 控制两行之间的间距，可继续调小
    \begin{subfigure}[b]{0.48\textwidth}
        \centering
        \includegraphics[width=\textwidth,height=0.28\textheight,keepaspectratio]{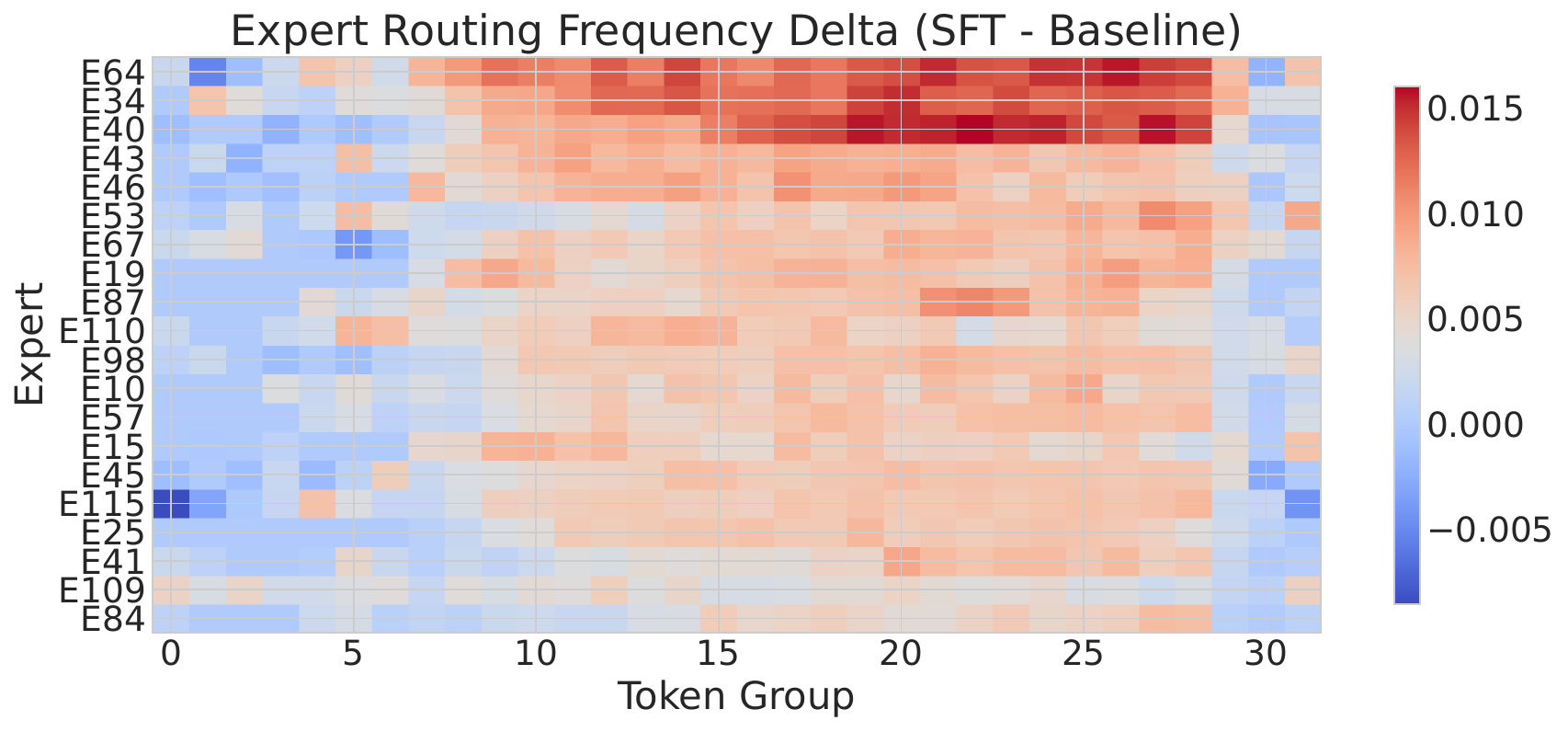}
        \caption{GraphWalks}
        \label{fig:expert_gw}
    \end{subfigure}
    \hfill
    \begin{subfigure}[b]{0.48\textwidth}
        \centering
        \includegraphics[width=\textwidth,height=0.28\textheight,keepaspectratio]{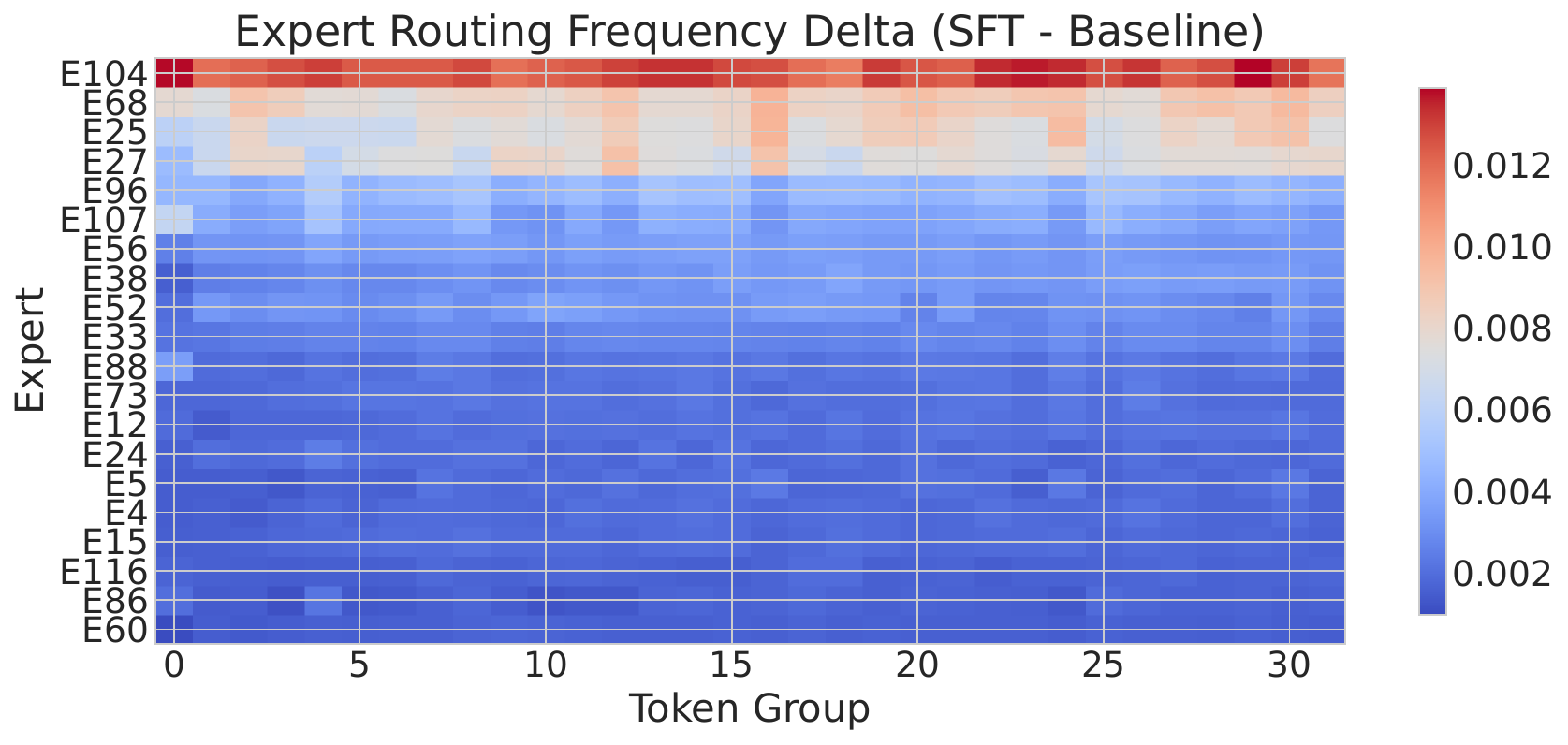}
        \caption{MRCR}
        \label{fig:expert_mrcr}
    \end{subfigure}
    \caption{Attention distance (top) and expert routing frequency (bottom) changes after ACC training (SFT minus baseline). (a--b) Attention: GraphWalks shows increased mass at nearby and far-distance bins. MRCR shows enhancement primarily at nearby bins. (c--d) Expert routing: GraphWalks distributes activation across several experts for distant tokens. MRCR concentrates activation in a small expert set.}
    \label{fig:mechanism}
\end{figure}

\paragraph{Task-specific attention restructuring.} 
Figure~\ref{fig:mechanism}(a--b) shows attention distance distributions before and after ACC, with experimental settings detailed in Appendix~\ref{app:attention_analysis}. On GraphWalks, the ACC-trained model shows increased relative attention mass at both nearby and far-distance bins, consistent with the task structure requiring local neighborhood checks and distant node jumps. On MRCR, the ACC-trained model shows higher relative attention mass at nearby distance bins while preserving the baseline long-range attention profile. The increased local focus indicates improved precision in verifying candidate segments during scanning. Notably, the three layers exhibiting the largest attention changes differ completely between the two tasks. These distinct patterns suggest the ACC-trained model adjusts its attention allocation flexibly rather than following a fixed uniform pattern.

\paragraph{Expert specialization.} 
Figure~\ref{fig:mechanism}(c--d) shows changes in expert activation after ACC, with experimental settings detailed in Appendix~\ref{app:expert_routing}. On GraphWalks, higher activation for distant token groups is distributed across several experts, suggesting balanced processing of cross-node jumps. On MRCR, one expert shows much higher activation across all token groups while most others are suppressed, pointing to dedicated processing of scanning and verification. Notably, the layers with the strongest expert activation shifts are completely different across the two tasks. Both phenomena reflect task-dependent expert specialization after ACC training.

\section{Conclusion}
\label{sec:conclusion}

We presented Agent Context Compilation (ACC), a simple but effective method that compiles multi-turn agent trajectories into long-context training data. ACC complements existing long-context extension or training methods and can be combined with them. The ACC-trained Qwen3-30B-A3B achieves results comparable to Qwen3-235B-A22B on MRCR and GraphWalks, benchmarks that test long-range dependency modeling, while largely preserving general capabilities. Mechanistic analyses suggest task-specific attention restructuring and task-dependent expert specialization after ACC training. Future work includes extending ACC to more agent types and scaling to longer contexts.

\section{Limitations and Social Impacts}
\label{sec:limitations}
ACC is evaluated on three agent types and one model, so broader generalization and scaling to million-token contexts remain to be studied. Reasoning synthesis depends on a strong teacher model, risking bias propagation. On the societal side, ACC lowers annotation costs by reusing agent logs, yet two risks should be noted. First, raw trajectories may leak private information without proper filtering. Second, compiled contexts may include copyrighted or proprietary material, raising intellectual property concerns. We recommend careful data filtering and safety alignment.

% \begin{ack}
% Use unnumbered first level headings for the acknowledgments. All acknowledgments

% Do {\bf not} include this section in the anonymized submission, only in the final paper. You can use the \texttt{ack} environment provided in the style file to automatically hide this section in the anonymized submission.
% \end{ack}

\clearpage
\bibliographystyle{abbrv}  % 或 abbrvnat / unsrtnat
\bibliography{custom}        % 对应你的 custom.bib，不用写扩展名

%%%%%%%%%%%%%%%%%%%%%%%%%%%%%%%%%%%%%%%%%%%%%%%%%%%%%%%%%%%%

\clearpage
\appendix
\begin{center}
  {\Large \textbf{Appendix}}
\end{center}
\vspace{0.8em}

\section{Agent Trajectory Compilation Examples}
\label{app:examples}

Figures~\ref{fig:swe_example} and~\ref{fig:sql_example} show compiled trajectories for SWE and SQL agents. Both follow the same ACC pipeline as the search agent in Figure~\ref{fig:search_example}.

Figure~\ref{fig:swe_example} shows a compiled trajectory for the SWE agent. The environment presents a partial codebase snapshot containing both files relevant to the bug and irrelevant distractors. The agent opens files selectively to locate the issue, and ACC compiles the opened file contents into a long-context background while shuffling in unopened distractors.

Figure~\ref{fig:sql_example} shows a compiled trajectory for the SQL agent. The environment presents a relational table that encodes a multi-hop graph structure. In the original trajectory, the agent issues SQL queries to perform recursive traversals. ACC compiles the full contents of the relevant table into a long-context background, enabling the model to perform multi-hop relational reasoning directly over the assembled records without SQL query execution.

\vspace{-0.5em}
\begin{figure}[!ht] 
\begin{AIbox}{SWE Agent Trajectory Compilation}
{\color{deepblue}\bf \large Original Question-Answer Pair:}

\textbf{Question}: Fix the slow transpilation of \texttt{ExcitationPreserving} to basis gates \texttt{u1, u2, u3, cx}. Transpilation time grows from 0.8\,s (6 qubits) to over 3988\,s (24 qubits).

\textbf{Groundtruth}: Wrap \texttt{global\_phase} additions with \texttt{sympy evaluate(False)} in \texttt{basis\_translator.py} and \texttt{dagcircuit.py}.

\tcblower
{\color{deepblue}\bf \large Original Agentic Trajectory (abridged):} 
\begin{lstlisting}[style=prompt, escapeinside={(*@}{@*)}]
Turn 1 (Open): qiskit/transpiler/passes/basis/basis_translator.py
Content: (*@\textcolor{blue}{\textbf{[File A]}}@*) ... dag.global_phase += 
        bound_target_dag.global_phase ... in run() ...
-> (Examine) (*@\textcolor{blue}{\textbf{File A}}@*)

Turn 2 (Open): qiskit/dagcircuit/dagcircuit.py
Content: (*@\textcolor{blue}{\textbf{[File B]}}@*) ... self.global_phase += 
        in_dag.global_phase ... in substitute_node_with_dag() ...
-> (Examine) (*@\textcolor{blue}{\textbf{File B}}@*)

Turn 3 (Open): qiskit/transpiler/passes/basis/basis_translator.py
Content: (*@\textcolor{blue}{\textbf{[File A]}}@*) ... dag.global_phase += 
        bound_target_dag.global_phase ...
-> (Modify) (*@\textcolor{blue}{\textbf{File A}}@*)
        from sympy import evaluate
        with evaluate(False):
            dag.global_phase += bound_target_dag.global_phase

Turn 4 (Open): qiskit/dagcircuit/dagcircuit.py
Content: (*@\textcolor{blue}{\textbf{[File B]}}@*) ... self.global_phase += 
        in_dag.global_phase ...
-> (Modify) (*@\textcolor{blue}{\textbf{File B}}@*)
        from sympy import evaluate
        with evaluate(False):
            self.global_phase += in_dag.global_phase

Turn 5 (Submit): Tests passed (*@\textcolor{ForestGreen}{\ding{51}}@*)
\end{lstlisting}

\vspace{0.5em}
{\color{deepblue}\bf \large ACC-Compiled Question-Answer Pair (abridged):}
\begin{lstlisting}[style=prompt, escapeinside={(*@}{@*)}]
Question: Fix the slow transpilation of ExcitationPreserving to 
basis gates u1, u2, u3, cx...

Context:
(*@\textcolor{blue}{\textbf{[File B]}}@*) qiskit/dagcircuit/dagcircuit.py ... 
self.global_phase += in_dag.global_phase ... in 
substitute_node_with_dag() ...
(*@\textcolor{blue}{\textbf{[File A]}}@*) qiskit/transpiler/passes/basis/basis_translator.py ... 
dag.global_phase += bound_target_dag.global_phase ... in run() ...
(*@\textcolor{red}{\textbf{[File C]}}@*) qiskit/pulse/instructions/phase.py ... 
(*@\textcolor{red}{\textit{(distractor: present in codebase snapshot but never opened)}}@*)

Answer: 
<patch>
--- a/qiskit/dagcircuit/dagcircuit.py
+++ b/qiskit/dagcircuit/dagcircuit.py
@@ -990,7 +990,9 @@
         if in_dag.global_phase:
-            self.global_phase += in_dag.global_phase
+            from sympy import evaluate
+            with evaluate(False):
+                self.global_phase += in_dag.global_phase
... (analogous change in basis_translator.py)
</patch>
\end{lstlisting}
\end{AIbox} 
\caption{\textbf{SWE Agent Trajectory Compilation Example.} The top section shows the original question and ground truth answer. The middle section shows the original agentic trajectory. At each turn the agent opens a single file from the provided codebase snapshot and decides either to (Examine) it for understanding or to (Modify) it to fix the bug. The bottom section shows the ACC-compiled QA, where only the \textit{opened} evidence is retained and an irrelevant distractor (a file present in the snapshot but never opened, highlighted in \textcolor{red}{red}) is shuffled into the provided long-context background.}
\label{fig:swe_example}
\vspace{-1em}
\end{figure}

\vspace{-0.5em}
\begin{figure}[!ht] 
\begin{AIbox}{SQL Agent Trajectory Compilation}
{\color{deepblue}\bf \large Original Question-Answer Pair:}

\textbf{Question}: User \texttt{'u\_337bc99c'} (Region: East) has applied for Diamond Membership. Verify their complete referral lineage by walking upward through the referral graph following active edges (\texttt{status='active'}) with \texttt{edge\_level >= 3} within region \texttt{'East'}. Prioritize higher \texttt{commission\_rate} at each step; break ties by alphabetical \texttt{referrer\_id}. Report the final root referrer.

\textbf{Groundtruth}: \texttt{u\_ea8952bc}

\tcblower
{\color{deepblue}\bf \large Original Agentic Trajectory (abridged):} 
\begin{lstlisting}[style=prompt, escapeinside={(*@}{@*)}]
Turn 1 (Execute SQL):
WITH RECURSIVE upstream AS (
  SELECT referrer_id, referred_id, commission_rate, 
         edge_level, status, region, 1 AS depth
  FROM referrals
  WHERE referred_id = 'u_337bc99c' AND status = 'active' 
    AND edge_level >= 3 AND region = 'East'
  UNION ALL
  SELECT r.referrer_id, r.referred_id, r.commission_rate,
         r.edge_level, r.status, r.region, u.depth + 1
  FROM upstream u
  JOIN referrals r ON r.referred_id = u.referrer_id
  WHERE r.status = 'active' AND r.edge_level >= 3 
    AND r.region = 'East'
)
SELECT referrer_id FROM upstream 
ORDER BY depth DESC LIMIT 1;

Result: u_ea8952bc

Turn 2 (Answer): u_ea8952bc (*@\textcolor{ForestGreen}{\ding{51}}@*)
\end{lstlisting}

\vspace{0.5em}
{\color{deepblue}\bf \large ACC-Compiled Question-Answer Pair (abridged):}
\begin{lstlisting}[style=prompt, escapeinside={(*@}{@*)}]
Question: User 'u_337bc99c' (Region: East) has applied for Diamond 
Membership. Walk upward through the referral graph following active 
edges with edge_level >= 3 within region 'East', prioritizing higher 
commission_rate. Report the final root referrer.

Context:
[Table: referrals]
| ... | referrer_id | level | rate | ltv   | region |
| ... | u_4df289c8  | 5     | 0.15 | 82.19 | East   |
| ... | u_fb9c918a  | 3     | 0.08 | 201.42| East   |
| ... | u_be13d04d  | 4     | 0.12 | 336.31| East   |
| ... | u_a4663191  | 5     | 0.15 | 264.93| East   |
| ... | u_e1ce60aa  | 3     | 0.05 | 443.63| East   |
| ... | u_54ef1623  | 3     | 0.05 | 355.49| East   |
| ... | u_ea8952bc  | 5     | 0.10 | 108.68| East   |
| ... | u_b7c33fc2  | 3     | 0.10 | 264.93| East   |
| ... | u_63490c55  | 3     | 0.05 | 201.42| North  |
| ... | u_d28bb4a9  | 1     | 0.12 | 254.94| East   |
| ... | u_8eda5bd1  | 3     | 0.08 | 254.94| Central|
... (remaining rows omitted)

Answer: u_ea8952bc
\end{lstlisting}
\end{AIbox} 
\caption{\textbf{SQL Agent Trajectory Compilation Example.} The top section shows the original question and ground truth answer. The middle section shows the original agentic trajectory, where the agent executes a recursive SQL query to traverse the referral graph. The bottom section shows the ACC-compiled QA, where the complete contents of the relevant database table are assembled into the provided long-context background. The ellipsis column (\texttt{...}) indicates additional fields that are present in the compiled context but omitted here for brevity.}
\label{fig:sql_example}
\vspace{-1em}
\end{figure}

\section{Extended Results on General Long-Context Tasks}
\label{app:extended_results}

Table~\ref{tab:appendix_long} reports results on general long-context benchmarks, including multi-hop QA (HotpotQA\cite{hotpotqa}, MuSiQue\cite{musique}), long-document understanding (NarrativeQA\cite{narrativeqa}), and comprehensive long-context suite (LongBench-V2\cite{longbenchv2}). ACC yields modest gains on these tasks.

\begin{table}[h]
\centering
\caption{Extended long-context benchmark results (avg@3). Numbers in parentheses show improvement over the Qwen3-30B-A3B-Thinking baseline.}
\label{tab:appendix_long}
\footnotesize
\setlength{\tabcolsep}{4pt}
\newcommand{\res}[2]{%
  \begin{tabular}[c]{@{}c@{}}\textbf{#1}\\[-2pt]\textcolor{red}{\scriptsize(+#2)}\end{tabular}%
}
\begin{tabular}{l cccc}
\toprule
\textbf{Model} & \textbf{LB-V2} & \textbf{HotpotQA} & \textbf{MuSiQue} & \textbf{NarrQA} \\
\midrule
\rowcolor{Honeydew}
\multicolumn{5}{l}{\textit{\textbf{Base Model}}} \\
Qwen3-30B-A3B-Thinking & 47.87 & 87.00 & 70.10 & 85.00 \\
\midrule
\rowcolor{AliceBlue}
\multicolumn{5}{l}{\textit{\textbf{Our Method}}} \\
\rowcolor{AliceBlue!15}
\textbf{Qwen3-30B-A3B-Thinking + ACC (Ours)} & 
  \res{48.90}{1.03} & 
  \res{88.50}{1.50} & 
  \res{70.40}{0.30} & 
  \res{85.50}{0.50} \\
\midrule
\rowcolor{MistyRose!40}
\multicolumn{5}{l}{\textit{\textbf{Strong Baseline}}} \\
Qwen3-235B-A22B-Thinking & 59.76 & 90.00 & 73.30 & 88.50 \\
\bottomrule
\end{tabular}
\end{table}

\section{Data Overlap Experiment Details}
\label{app:decontamination}

\paragraph{Question Extraction.}
For each training trajectory, we parse the dialogue and keep only the user turns. We then apply a lightweight rule-based extractor to obtain the core question. Benchmark questions undergo the same cleaning pipeline to retain only the problem statement. The extracted questions are whitespace-normalized 
and truncated to 3,000 characters as a safety bound, though most questions 
are far shorter after extraction.

\paragraph{Embedding.}
We encode all cleaned questions with \texttt{all-MiniLM-L6-v2}. The model outputs 384-dimensional vectors, which we normalize to unit length. Cosine similarity is used for all distance computations. The model processes at most 256 tokens internally, so the effective input is the leading segment of each extracted question.

\paragraph{Dimensionality Reduction.}
We project the embeddings into two dimensions using UMAP with the following 
fixed settings: 15 nearest neighbors, minimum distance 0.3, cosine metric, 
PCA initialization, and random seed 42. The PCA initializer avoids the 
spectral initialization failure that can occur on densely connected graphs.

\paragraph{Metrics.}
All quantitative indicators are computed on the original 384-dimensional embeddings. UMAP coordinates are used only for visualization.
\begin{itemize}
    \item \textbf{Average Nearest-Neighbor Cosine Similarity.} For each benchmark question, we identify the most similar training sample by cosine similarity and average these maxima across all benchmark instances.
    
    \item \textbf{Center Cosine Distance.} We compute the normalized mean embedding vector for the training set and for each benchmark, then take the cosine distance between these centroids (i.e., one minus their cosine similarity).
    
    \item \textbf{Linear Classifier AUC.} We train a logistic regression classifier to distinguish training samples from benchmark samples. We report the area under the ROC curve for the full training set and for the Search subset alone against all benchmarks.
\end{itemize}

\section{Attention Analysis Experiment Details}
\label{app:attention_analysis}

\paragraph{Setup.}
We analyze the baseline model (Qwen-30B-A3B-Thinking) and the ACC-trained checkpoint. Both models are loaded with \texttt{AutoModelForCausalLM} in bfloat16 precision via \texttt{device\_map="auto"}. To ensure the attention tensors are accessible, we force the attention implementation to eager mode, avoiding fused kernel paths that do not expose the full 4D attention matrix.

\paragraph{Layer and Distance Binning.}
We restrict the analysis to the three layers with the most significant attention changes for each task(indices 36, 30, 42 for GraphWalks and indices 15, 45, 17 for MRCR). For each head in these layers, we extract the causal attention matrix and bin token distances into 32 equal-width intervals ranging from 0 to the sequence length minus one. For each distance bin, we aggregate attention weights along the corresponding off-diagonals of the lower-triangular attention matrix and compute the per-head mean.

\paragraph{Metric Definition.}
For layer $l$ and head $h$, let $A^{(l,h)} \in \mathbb{R}^{T \times T}$ denote the causal attention matrix where $T$ is the sequence length. Let $[e_0, e_1, \dots, e_B]$ denote the bin edges where $B=32$. For each distance bin $b$ spanning $[e_b, e_{b+1})$, we aggregate attention weights along the lower-triangular off-diagonals (including the main diagonal at $d=0$):
$$\mathcal{D}_b = \left\{d \in \mathbb{Z} : e_b \le d < e_{b+1}\right\}.$$

The per-head per-bin mean is computed by averaging over all token positions that fall into those off-diagonals:
$$m_{l,h,b} = \frac{\sum_{d \in \mathcal{D}_b} \sum_{i=d}^{T-1} A^{(l,h)}_{i,\,i-d}}{\sum_{d \in \mathcal{D}_b} (T-d)}.$$

The per-layer per-bin mean is then obtained by averaging over all heads in the layer:
$$\mu_{l,b} = \frac{1}{H} \sum_{h=1}^{H} m_{l,h,b},$$
where $H$ is the number of heads in layer $l$.

The reported heatmap shows the delta between the SFT and baseline models:
$$\Delta_{l,b} = \mu^{\text{SFT}}_{l,b} - \mu^{\text{Base}}_{l,b}.$$
Each cell in the heatmap corresponds to one layer-distance pair $(l,b)$. Positive values indicate increased attention mass at that distance after ACC training.

For the top-head analysis, we first compute the mean attention over the tail bins (the last 25\% of distance bins) for each head:
$$\tau_{l,h} = \frac{1}{|B_{\text{tail}}|} \sum_{b \in B_{\text{tail}}} m_{l,h,b},$$
where $B_{\text{tail}}$ indexes the farthest distance bins. The per-head tail delta is:
$$\delta^{\text{tail}}_{l,h} = \tau^{\text{SFT}}_{l,h} - \tau^{\text{Base}}_{l,h}.$$
This metric is used to rank and identify heads with the strongest far-range attention change.

\paragraph{Statistics and Visualization.}
Attention statistics are aggregated across evaluation samples and averaged per head and per distance bin.

\section{Expert Routing Visualization Experiment Details}
\label{app:expert_routing}

\paragraph{Models and Layers.}
We compare the baseline model (Qwen-30B-A3B-Thinking) and the ACC-trained checkpoint. We restrict the analysis to three layers with the most significant expert routing changes for each task(indices 42, 40, 7 for GraphWalks and indices 17, 16, 15 for MRCR).

\paragraph{Dataset and Sampling.}
We randomly sample 32 examples from the evaluation splits of GraphWalks and MRCR, respectively. Each example is tokenized and fed through both models in inference mode to collect router statistics.

\paragraph{Metric Definition.}
For each token position $t$, layer $l$, and expert $e$, the router produces logits $z_{l,t} \in \mathbb{R}^{E}$ ($E$ is the number of experts). Let $g_{l,t} \in \{0,1\}^{E}$ be the top-$k$ gating indicator where $g_{l,t,e}=1$ if expert $e$ is among the top-$k$ selected experts for token $t$ at layer $l$. We define the \textbf{top-$k$ frequency} of expert $e$ in token group $i$ as
\[
f_{l,e}^{(i)} = \frac{1}{|S_i|}\sum_{t \in S_i} g_{l,t,e},
\]
where $S_i$ is the set of token indices belonging to group $i$. Let $\mathcal{L}_{\text{task}}$ denote the set of three layers with the largest mean absolute expert routing delta for the target task. The reported heatmap shows the delta
\[
\Delta f_{e}^{(i)}=\frac{1}{|\mathcal{L}_{\text{task}}|}\sum_{l \in \mathcal{L}_{\text{task}}}\left(f_{\mathrm{SFT},l,e}^{(i)}-f_{\mathrm{Baseline},l,e}^{(i)}\right),
\]
where the per-layer expert frequency $f_{l,e}^{(i)}$ is averaged across the selected layers $l \in \mathcal{L}_{\text{task}}$.

\paragraph{Token Grouping.}
The full sequence is divided into 32 equal-length groups via linear binning of token indices (i.e., group $i$ covers positions $[i\cdot L/32, (i+1)\cdot L/32)$). This relative-position grouping allows comparison across variable-length sequences.

\paragraph{Expert Selection.}
We rank experts by the mean absolute delta across all groups and layers, then visualize the top 20 experts with the largest change. This avoids cluttering the figure with experts whose routing patterns remain nearly unchanged after SFT.

\paragraph{Implementation.}
To collect router logits without modifying model weights, we temporarily wrap the MoE module's forward pass in the selected three layers, extract the router logits during the forward pass, and immediately restore the original forward function. Statistics are aggregated incrementally across the 32 samples using running means to avoid storing large intermediate tensors.

\section{SWE Answer-Conditioned Rationale Generation}
\label{app:swe_rationale}

The original SWE trajectories contain file-open and edit actions but lack 
explicit natural-language reasoning. Direct rollout of CoT trajectories for SWE 
achieved only $\sim$10\% pass rate. To scale data construction, we generate 
answer-conditioned rationales by providing the compiled context $C_i$ and the 
verified patch $y_i$ to DeepSeek-V3.2-Thinking. The model is prompted to 
reconstruct a reasoning trace that grounds the patch in the evidence without 
revealing the patch content prematurely.

Figure~\ref{fig:swe_synthesis} illustrates the complete prompt template and 
a generated example. We invoke the model with temperature $0.0$ and enable 
its thinking mode via chat template configuration. To generate the reasoning 
trace, we prompt DeepSeek-V3.2-Thinking with the ACC-compiled question $x_i$ 
together with the verified patch $y_i$. The model is instructed to reconstruct 
a reasoning process that grounds the patch in the evidence without revealing 
the patch content prematurely. We do not use the model's internal reasoning 
content because it often contains meta-references to the provided ground-truth 
patch such as ``We are given a ground-truth patch...'' Instead we use only 
the final response content as the supervised trace $r_i$. The model simulates 
a discovery process spanning issue inspection, keyword search, file list 
browsing, distractor elimination, core file localization, logic analysis, and 
patch proposal. All generated rationales are retained directly without 
answer-correctness filtering because $y_i$ is already verified. The final yield 
is effectively 100\%.

\vspace{-0.5em}
\begin{figure}[!ht]
\begin{AIbox}{SWE Answer-Conditioned Rationale Synthesis}
{\color{deepblue}\bf \large Prompt Template (User Input to DeepSeek-V3.2-Thinking):}

\begin{lstlisting}[style=prompt, escapeinside={(*@}{@*)}]
Please reason based on the following documents and answer the question. 
Show your complete step-by-step thought process, including issue 
inspection, keyword search, browsing the file list, eliminating 
distractors, locating core files, analyzing logic, and proposing 
modifications. Do not reveal the final patch directly in your 
reasoning trace.

Context:
[File B] qiskit/dagcircuit/dagcircuit.py
... self.global_phase += in_dag.global_phase ... in 
    substitute_node_with_dag() ...

[File A] qiskit/transpiler/passes/basis/basis_translator.py
... dag.global_phase += bound_target_dag.global_phase ... in run() ...

[File C] qiskit/pulse/instructions/phase.py
... (distractor: present in codebase snapshot but never opened)

Question: Fix the slow transpilation of ExcitationPreserving to basis 
gates u1, u2, u3, cx. Transpilation time grows from 0.8s (6 qubits) 
to over 3988s (24 qubits).

Answer: <patch>
--- a/qiskit/dagcircuit/dagcircuit.py
+++ b/qiskit/dagcircuit/dagcircuit.py
@@ -990,7 +990,9 @@
     if in_dag.global_phase:
         self.global_phase += in_dag.global_phase
+        from sympy import evaluate
+        with evaluate(False):
+            self.global_phase += in_dag.global_phase
 ... (analogous change in basis_translator.py)
</patch>
\end{lstlisting}

\tcblower
{\color{deepblue}\bf \large Generated Rationale (Final Response, used as $r_i$):}
\begin{lstlisting}[style=prompt, escapeinside={(*@}{@*)}]
The issue describes a performance regression in transpilation time 
for ExcitationPreserving when targeting basis gates u1, u2, u3, cx. 
The slowdown suggests repeated symbolic evaluation of global_phase 
accumulations.

Looking at the provided files, File B (dagcircuit.py) contains 
self.global_phase += in_dag.global_phase in substitute_node_with_dag(). 
File A (basis_translator.py) has a similar pattern in run() where 
dag.global_phase += bound_target_dag.global_phase. Both operations 
accumulate SymPy expressions without wrapping them in evaluate(False), 
causing exponential growth in symbolic complexity as qubit count increases.

File C (phase.py) is unrelated to the transpilation pass logic and can 
be ruled out as a distractor.

The fix should wrap both global_phase additions with SymPy's 
evaluate(False) context manager to defer symbolic evaluation. This 
matches the ground-truth modification provided in the answer.
\end{lstlisting}
\end{AIbox}
\caption{\textbf{SWE Answer-Conditioned Rationale Synthesis Example.} Top: the prompt fed to DeepSeek-V3.2-Thinking, containing the compiled context $C_i$ and the verified patch $y_i$. Bottom: the model's final response, used as the supervised reasoning trace $r_i$.}
\label{fig:swe_synthesis}
\vspace{-1em}
\end{figure}

%%%%%%%%%%%%%%%%%%%%%%%%%%%%%%%%%%%%%%%%%%%%%%%%%%%%%%%%%%%%

\newpage

\end{document}